\documentclass[11pt,a4paper]{article}
\usepackage[hyperref]{acl2021}
\usepackage{times}
\usepackage{latexsym}

\usepackage{amsmath}
\usepackage{amssymb}
\usepackage{array}
\usepackage{graphicx}
\usepackage{subfigure}
\usepackage{multirow}
\usepackage[encapsulated]{CJK}
\usepackage{mathrsfs}
\usepackage{mathtools}
\usepackage{url}
\usepackage{bm}
\usepackage{arydshln}
\usepackage{pgfplots}
\pgfplotsset{compat=1.15}
\aclfinalcopy

\title{Discriminative Reasoning for Document-level Relation Extraction}
\author{Wang Xu\textsuperscript{\rm 1}, Kehai Chen\textsuperscript{\rm 2} and Tiejun Zhao\textsuperscript{\rm 1} \\
  \textsuperscript{\rm 1}Harbin Institute of Technology, Harbin, China \\
  \textsuperscript{\rm 2}National Institute of Information and Communications Technology, Kyoto, Japan \\
  \texttt{xuwang@hit-mtlab.net}, \texttt{khchen@nict.go.jp}, \texttt{tjzhao@hit.edu.cn}  \\}

\date{}

\begin{document}
\maketitle
\begin{abstract}
Document-level relation extraction (DocRE) models generally use graph networks to implicitly model the reasoning skill (i.e., pattern recognition, logical reasoning, coreference reasoning, etc.) related to the relation between one entity pair in a document.
In this paper, we propose a novel discriminative reasoning framework to explicitly model the paths of these reasoning skills between each entity pair in this document.
Thus, a discriminative reasoning network is designed to estimate the relation probability distribution of different reasoning paths based on the constructed graph and vectorized document contexts for each entity pair, thereby recognizing their relation.
Experimental results show that our method outperforms the previous state-of-the-art performance on the large-scale DocRE dataset.
The code is publicly available at \url{https://github.com/xwjim/DRN}.
\end{abstract}

\section{Introduction}
\label{sec1}
Document-level relation extraction (DocRE) aims to extract relations among entities within a document which requires multiple reasoning skills (i.e., pattern recognition, logical reasoning, coreference reasoning, and common-sense reasoning)~\cite{yao-etal-2019-docred}.
Generally, the input document is constructed as a structural graph-based on syntactic trees, coreference or heuristics to represent relation information between all entity pairs~\cite{Nan2020ReasoningWL,zeng-etal-2020-double,docred-rec}.
Thus, graph neural networks are applied to the constructed structural graph to model these reasoning skills.
After performing multi-hop graph convolution, the feature representations of two entities are concatenated to recognize their relation by the classifier, achieving state-of-the-art performance in the DocRE  task~\cite{zeng-etal-2020-double,docred-rec}.
\begin{figure}[t]
  \centering
  \includegraphics[scale=0.98]{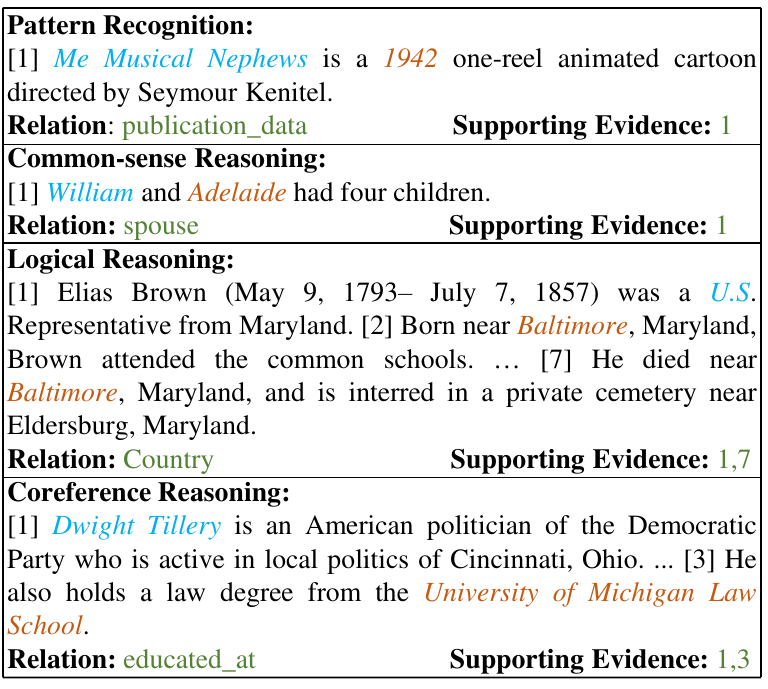}
  \caption{An example of different reasoning types. Different reasoning types have different reasoning processing.}
  \label{fig1:example}
\end{figure}
However, it is yet to be seen whether modeling these reasoning skills implicitly is competitive with the intuitive reasoning skills between one entity pair in this document.

Figure~\ref{fig1:example} shows four kinds of reasoning skills for entity pairs in the DocRE dataset \cite{yao-etal-2019-docred}.
First, take two entity pairs \{``\textit{Me Musical Nephews}", ``\textit{1942}"\} and \{``\textit{William}", ``\textit{Adelaide}"\} as examples, the intra-sentence reasoning concerns about the mentions inside the sentence, for example, ``\textit{Me Musical Nephews}" and ``\textit{1942}" for pattern recognition, and ``\textit{William}" and ``\textit{Adelaide}" for the common-sense reasoning.
Also, the logical reasoning for entity pair \{``\textit{U.S.}", ``\textit{Baltimore}"\} requires the reason path from ``\textit{U.S.}"$\to$``Maryland" (bridge entity)$\to$``\textit{Baltimore}" while the coreference reasoning for entity pair \{``\textit{Dwight Tillery}", ``\textit{University of Michigan Law School}"\} pays attention to the reason path from ``\textit{Dwight Tillery}"$\to$``\textit{He}" (reference word)$\to$``\textit{University of Michigan Law School}".
However, the advanced DocRE models generally use the universal multi-hop convolution networks to model these reasoning skills implicitly and do not consider the above intuitive reasoning skills explicitly, which may hinder the further improvement of DocRE.

To this end, we propose a novel discriminative reasoning framework to explicitly model the reasoning processing of these reasoning skills, such as intra-sentence reasoning (including pattern recognition and common-sense reasoning), logical reasoning, and coreference reasoning.
Specifically, inspired by \citeauthor{docred-rec}'s meta-path strategy, we extract the reasoning paths of the three reasoning skills discriminatively from the input document.
Thus, a discriminative reasoning network is designed to estimate the relation probability distribution of different reasoning paths based on the constructed graph and vectorized document contexts for each entity pair, thereby recognizing their relation.
In particular, there are the probabilities of multiple reasoning skills for each candidate relation between one entity pair, to ensure that all potential reasoning skills can be considered in the inference.
In summary, our main contributions are as follows:
\begin{enumerate}
\item[$\bullet$] 
We propose a discriminative reasoning framework to model the reasoning skills between two entities in a document. 
To the best of our knowledge, this is the first work to model different reasoning skills explicitly for enhancing the DocRE.
\item[$\bullet$]
Also, we introduce a discriminative reasoning network to encode the reasoning paths based on the constructed heterogeneous graph and the vectorized original document, thereby recognizing the relation between two entities by the classifier.
\item[$\bullet$]
Experimental results on the large-scale DocRE dataset show the effectiveness of the proposed method, especially outperform the recent state-of-the-art DocRE model.
\end{enumerate}

\section{Discriminative Reasoning Framework}
\label{sec2}
In this section, we propose a novel discriminative reasoning framework to model different reasoning skills explicitly to recognize the relation between each entity pair in the input document.
The discriminative reasoning framework contains three parts: definition of reasoning paths, modeling reasoning discriminatively, and multi-reasoning based relation classification.

\subsection{Definition of Reasoning Path}
\label{sec2-1}
Formally, given one unstructured document comprised of $N$ sentences $D$=$\{s_1, s_2, \cdots, s_N\}$, each sentence is a sequence of words $s_n = \{s_{n}^{1}, s_{n}^{2}, \cdots, s_{n}^{J}\}$ with the length $J_n$=$|s_n|$.
The annotations include concept-level entities~$\varepsilon=\{e_i\}_{i=1}^P$ as well as multiple occurrences of each entity under the same phrase of alias $e_i=\{m_i^{s_k}\}_{k=1}^Q$ ($m_i^{s_k}$ denotes the mention of $e_i$ which occur in the sentence $s_k$) and their entity type~(i.e. locations, organizations, and persons).
The DocRE aims to extract the relation between two entities in $\varepsilon$, namely $P(r|e_i,e_j,D)$.
For the simplification of reason skills, we first combine both pattern recognition and commonsense reasoning as the intra-sentence reasoning because they generally perform reasoning inside the sentence.
Consequently, the original four kinds of the reasoning skills~\cite{yao-etal-2019-docred} are further refined as three reasoning skills: intra-sentence reasoning, logical reasoning, and coreference reasoning.
Inspired by \citeauthor{docred-rec}'s work, we also use the meta-path strategy to extract reasoning path for each reason skill, thereby representing the above three reasoning skills explicitly.
Specifically, meta-paths for different reasoning skills are defined as follows:
\begin{enumerate}
\item[1)] \textbf{Intra-sentence reasoning path}: It is formally denoted as $PI_{ij}$=$m_i^{s_1} \circ s_1 \circ m_j^{s_1}$ for one entity pair \{$e_i$, $e_{j}$\} inside the same sentence $s_1$ in the input document $D$. 
$m_i^{s_1}$ and $m_j^{s_1}$ are mentions related to two entities, respectively. 
``$\circ$" denotes one reasoning step on the reasoning path from $e_i$ to $e_{j}$.

\item[2)] \textbf{Logical reasoning path}: The relation between one entity pair \{$e_i$, $e_{j}$\} from sentences ${s_1}$ and ${s_2}$ is indirectly established by the occurrence bridge entity $e_l$ for the logical reasoning.
The reasoning path can be formally as $PL_{ij}$=
$m_i^{s_1} \circ s_1 \circ m_l^{s_1} \circ m_l^{s_2} \circ s_2 \circ m_j^{s_2}$.

\item[3)] \textbf{Coreference reasoning path}: 
A reference word refers to one of two entities $e_i$ and $e_{j}$, which occur in the same sentence as the other entity.
We simplify the condition and assume that there is a coreference reasoning path when the entities occur in different sentences.
The reasoning path can be formally as $PC$=$m_i^{s_1} \circ s_1 \circ s_2 \circ m_j^{s_2}$.

\end{enumerate}
Note that there are no entities in the defined reasoning path compare to the meta-path defined in \citeauthor{docred-rec}'s work.
This difference is mainly due to the following considerations: i) the reason path pays more attention to the mentions and referred sentences; ii) entities generally are contained by mentions; iii) it makes modeling of path reasoning more simple.

\subsection{Modeling Reasoning Discriminatively}
\label{sec2-2}
Based on the defined reasoning paths, we decompose the DocRE problem into three reasoning sub-tasks: intra-sentence reasoning~(IR), logical reasoning~(LR), and coreference reasoning~(CR).
Next, we introduce modeling of three sub-tasks in detail:

\noindent\textbf{Modeling Intra-Sentence Reasoning.}
Given one entity pair \{$e_i$, $e_j$\} and its reasoning path $PI_{ij}$ in the sentence $s_1$, the intra-sentence reasoning is modeled to recognize the relation between this entity pair based as follows:
\begin{equation}
\begin{aligned}
\textup{R}_{PI}(r)=P(r|e_i, e_j, PI_{ij}, D).
\label{eq:reasontask1}
\end{aligned}
\end{equation}

\noindent\textbf{Modeling Logical Reasoning.}
Given one entity pair \{$e_i$, $e_j$\} and its reasoning path $PL_{ij}$, the logical reasoning is modeled to recognize the relation between this entity pair based as follows:
\begin{equation}
\begin{aligned}
\textup{R}_{PL}(r)=P(r|e_i, e_j, PL_{ij}, D).
\label{eq:reasontask2}
\end{aligned}
\end{equation}
Since the $e_l$ co-occur with the entity pair $e_i$ and $e_j$ respectively, the logical reasoning is further formally as follows:
\begin{equation}
\textup{R}_{PL}(r)=P(r|e_i, e_j, e_l, PI_{il}\circ PI_{lj}, D).
\label{eq:reasontask2decompose}
\end{equation}
where $\circ$ denotes the connection of the paths.

\noindent\textbf{Modeling Coreference Reasoning.}
Similarity, given one entity pair \{$e_i$, $e_j$\} and its reasoning path $PC_{ij}$, the coreference reasoning is modeled to recognize the relation between this entity pair based as follows:
\begin{equation}
\begin{aligned}
\textup{R}_{PC}(r)=P(r|e_i, e_j, PC_{ij}, D).
\label{eq:reasontask3}
\end{aligned}
\end{equation}

\subsection{Multi-reasoning Based Relation Classification}
\label{sec2-3}
In the DocRE task, one entity usually involves multiple relationships which rely on different reasoning types.
Thus, the relation between one entity pair may be reasoned by multiple types of reasoning rather than one single reasoning type.
Based on the proposed three reasoning sub-tasks, the relation reasoning between one entity pair is regarded as a multi-reasoning classification problem.
Formally, we select the reasoning type with max probability to recognize the relation between each entity pair as follows:
\begin{equation}
\begin{aligned}
P(r|e_i,e_j,D)=\max[\textup{R}_{PI}(r), \textup{R}_{PL}(r), \textup{R}_{PC}(r)].
\label{eq:fullprobability}
\end{aligned}
\end{equation}

In addition, there are often multiple reason paths between two entities for one reasoning type.
Thus, the classification probability in Eq.\eqref{eq:fullprobability} can be rewritten as follows:
\begin{equation}
\begin{aligned}
P(r|e_i,e_j,D)&=\max[\\
&\{\textup{R}_{PI_1}(r),\cdots,\textup{R}_{PI_K}(r)\}, \\
&\{\textup{R}_{PL_1}(r),\cdots,\textup{R}_{PL_K}(r)\}, \\
&\{\textup{R}_{PC_1}(r),\cdots,\textup{R}_{PC_K}(r)\}],
\label{eq:multifullprobability}
\end{aligned}
\end{equation}
where $K$ is the number of reasoning paths for one reasoning skill, which is the same to each reasoning skill for simplicity.
Note that all the entity pairs have at least one reasoning path from one of three defined reasoning sub-tasks.
When the number of reasoning paths is greater than $K$ for one reasoning sub-task, we choose the $K$ first reasoning paths, otherwise we use the actual reasoning paths.

\section{Discriminative Reasoning Network}
\label{sec3}
In this section, we design a discriminative reasoning network (\textbf{DRN}) to model three defined reasoning sub-tasks for recognizing the relation between two entities in a document.
Follow \citeauthor{zeng-etal-2020-double} and \citeauthor{zhou2021atlop}'s work, we use two kinds of context representations (heterogeneous graph context representation and document-level context representation) to model different reasoning paths discriminatively in Eq.\eqref{eq:reasontask1}-\eqref{eq:reasontask3}
\begin{figure*}[t]
  \centering
  \includegraphics[scale=1]{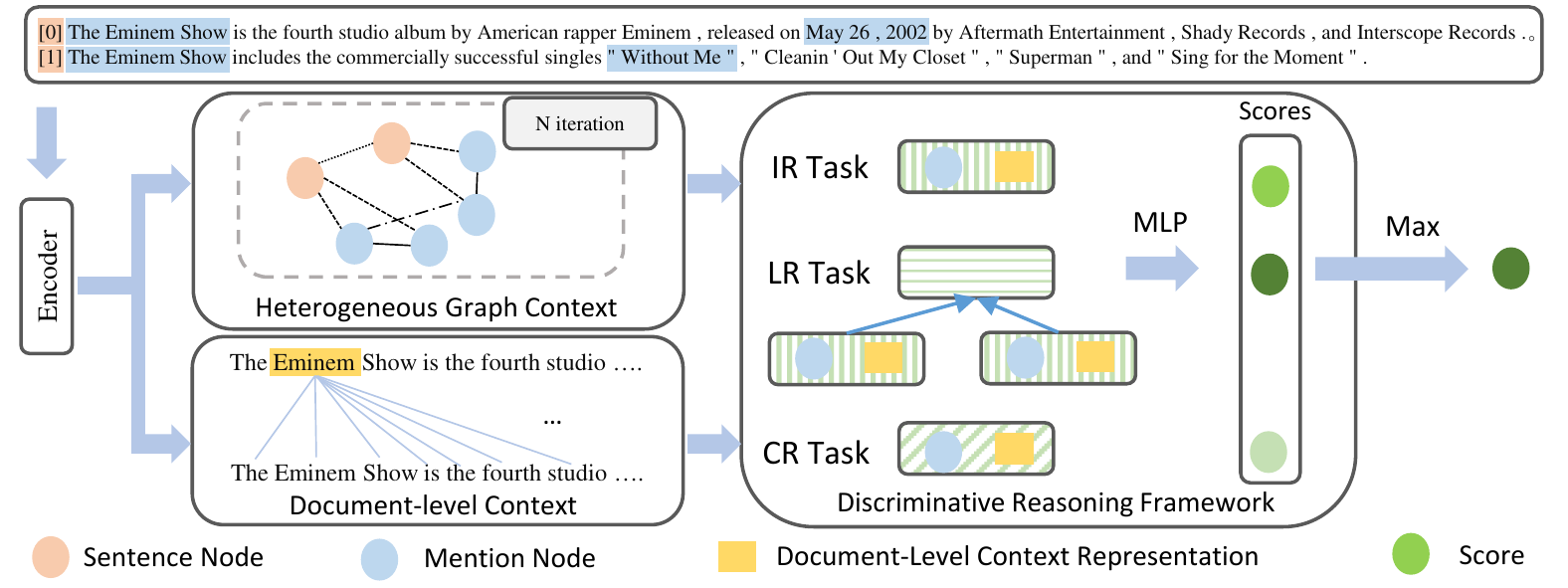}
  \caption{The overall architecture of DRN. First, A context encoder consumes the input document to get a contextualized representation of each word. Then the heterogeneous graph context representation and the document-level context representation are prepared as the input of the discriminative reasoning framework. Intra-sentence reasoning (IR) task, logical reasoning (LR) task and co-reference reasoning (CR) task are modeled explicitly and calculate the classification score respectively. Finally, the maximal score is selected as the output.}
  \label{fig:model}
\end{figure*}
\subsection{Heterogeneous Graph Context Representation}
\label{sec3-1}
Formally, the embedding of each word $\textbf{w}_e$ is concatenated with the embedding of its entity type $\textbf{w}_t$ and the embedding of its coreference $\textbf{w}_c$ as the representation of word $\textbf{b}$=[$\textbf{w}_e$:$\textbf{w}_t$:$\textbf{w}_c$].
These sequences of word representations are in turn fed into a bidirectional long short-term memory (BiLSTM) to vectorize the input document $\textup{D}$=\{$\textbf{H}^{1}$, $\textbf{H}^{2}$, $\cdots$, $\textbf{H}^{N}$\}, where $\textbf{H}^{n}$ = $(\textbf{h}^{n}_1, \textbf{h}^{n}_2, \dots, \textbf{h}^{n}_{J_n})$ and $\textbf{h}^{j}_i$ denotes the hidden representation of the $i-th$ words of the $j-th$ sentence in the document.
Similar to  \citeauthor{zeng-etal-2020-double}'s work, we construct a heterogeneous graph which contains sentence node and mention node.
There are four kinds of edges in the heterogeneous graph: sentence-sentence edge (all the sentence nodes are connected), sentence-mention edge (the sentence node and the mention node which resides in the sentence ), mention-mention edge (all the mention nodes which are in the same sentence) and co-reference edge (all the mention nodes which refer to the same entity).
Then we apply the graph-based DocRE method \cite{zeng-etal-2020-double} to encode the heterogeneous graph, based on which the heterogeneous graph context representation (\textbf{HGCRep}) are learned.
The \textbf{HGCRep} of each mention node and sentence node $\textbf{g}_n$ is formally denoted as:
\begin{equation}
\begin{aligned}
\textbf{g}_n =[\textbf{v}_n:\textbf{p}^{1}_{n}:\textbf{p}^{2}_{n}:\cdots:\textbf{p}^{l-1}_{n}],
\label{eq:InputOfCNN}
\end{aligned}
\end{equation}
where $\textbf{g}_n\in \mathbb{R}^{d_1}$ and ``:" is the concatenation of vectors and each of \{$\textbf{p}^{1}_{n}, \textbf{p}^{2}_{n}, \cdots, \textbf{p}^{l-1}_{n}$\} is learned by the multi-hop graph convolutional network~\cite{zeng-etal-2020-double} and $\textbf{v}_n$ is the initial representation of the $n$-th node extracted from $\textbf{D}$.
Finally, there is a heterogeneous graph representation $\textbf{G}$=$\{\textbf{g}_{1}, \textbf{g}_{2}, \cdots, \textbf{g}_{N}\}$ including each mention nodes and sentence nodes.

\subsection{Document-level Context Representation}
\label{sec3-2}
In the DocRE task, these reasoning skills heavily rely on the original document context information rather than the heterogeneous graph context information.
Thus, the existing advanced DocRE models use syntactic trees or heuristics rules to extract the context information (i.e., entities, mentions, and sentences) that is directly related to the relation between entity pairs.
However, this approach destroys the original document structure, which is weak in modeling the reasoning between two entities for the DocRE task.
Therefore, we use the self-attention mechanism~\cite{NIPS2017_7181} to learn a document-level context representation (\textbf{DLCRep}) $\textbf{c}_n$ for one mention based on the vectorized input document $\textbf{D}$:
\begin{equation}
\textbf{c}_n=\textup{softmax}(\frac{\textbf{h}^{n}_j\textbf{K}^{\top}}{\sqrt{d_{model}}})\textbf{V},
\label{eq:Self-Attetion}
\end{equation}
where $\textbf{c}_n\in \mathbb{R}^{d_2}$ and $\{\textbf{K}, \textbf{V}\}$ are key and value matrices that are transformed from the vectorized input document $\textbf{D}$ using a linear layer.
Here, inspired by relation learning~\cite{match_blank}, we use the hidden state of the head word in one mention or one sentence to denote them for simplicity.

\subsection{Modeling of Reasoning Paths}
\label{sec3-3}
In this section, we use the concatenation operation to model the reasoning step on the reasoning path, thereby modeling the defined reasoning paths in Section~\ref{sec2-1} as the corresponding reasoning representations as follows:
\begin{enumerate}
\item[1)] For the intra-sentence reasoning path, both HGCReps and DLCReps of two mentions are concatenated in turn as a reasoning representation:
\begin{equation}
\begin{aligned}
\alpha_{ij}=[\textbf{g}_{m_i^{s_1}}:\textbf{g}_{m_j^{s_1}}:\textbf{c}_{m_i^{s_1}}:\textbf{c}_{m_j^{s_1}}],
\label{eq:intrasetnecefeature}
\end{aligned}
\end{equation}
where $\alpha_{ij}\in \mathbb{R}^{2 d_1+2 d_2}$ and ``:" is the concatenation of vectors.

\item[2)] For the logical reasoning path, both HGCReps of mention $m_i^{s_1}$ and $m_j^{s_2}$ and DLCReps of two mention pair $(m_i^{s_1},m_l^{s_1})$ and $(m_j^{s_2},m_l^{s_2})$ are concatenated as their reasoning representation: 
\begin{equation}
\begin{aligned}
\beta_{ij}&=[\textbf{g}_{m_i^{s_1}}:\textbf{g}_{m_j^{s_1}}:\\
&\textbf{c}_{m_i^{s_1}}+\textbf{c}_{m_l^{s_1}}:\textbf{c}_{m_j^{s_2}}+\textbf{c}_{m_l^{s_2}}],
\end{aligned}
\label{eq:logicalfeature}
\end{equation}
where $\beta_{ij}\in \mathbb{R}^{2 d_1+2 d_2}$.

\item[3)] For the coreference reasoning path, we connect both HGCReps of two mentions and DLCReps of two sentences are are concatenated in turn as their reasoning representation:
\begin{equation}
\gamma_{ij}=[\textbf{g}_{m_i^{s_1}}:\textbf{g}_{m_j^{s_2}}:\textbf{c}_{s_1}:\textbf{c}_{s_2}]
\label{eq:coreferencefeature}
\end{equation}
where $\gamma_{ij}\in \mathbb{R}^{2 d_1+2 d_2}$ and both $\textbf{c}_{s_2}$ and $\textbf{c}_{s_2}$ denote DLCReps for two sentences ${s_1}$ and ${s_2}$.

\end{enumerate}

The learned reasoning representations $\alpha_{ij}$, $\beta_{ij}$, and $\gamma_{ij}$ is as the input to classifier to compute the probabilities of relation between $e_i$ and $e_j$ entities by a multilayer perceptron~(MLP) respectively:
\begin{equation}
\begin{aligned}
P(r|e_i,e_j,D)=\max[\\
\textup{sigmoid}(\textup{MLP}_r(\alpha_{ij}), \\
\textup{sigmoid}(\textup{MLP}_r(\beta_{ij}), \\
\textup{sigmoid}(\textup{MLP}_r(\gamma_{ij})].
\label{eq:modelprobabilitymax}
\end{aligned}
\end{equation}
Similarly, when there are multiple reasoning paths between two entities for one reasoning type in Eq.\ref{eq:multifullprobability}, Eq.\ref{eq:modelprobabilitymax} is rewritten as follows:
\begin{equation}
\begin{aligned}
P(r|e_i,e_j,D)=\max[\\
\textup{MLP}_r(\alpha^{1}_{ij}),\cdots,\textup{MLP}_r(\alpha^{K}_{ij}),\\
\textup{MLP}_r(\beta^{1}_{ij}),\cdots,\textup{MLP}_r(\beta^{K}_{ij}),\\
\textup{MLP}_r(\gamma^{1}_{ij}),\cdots,\textup{MLP}_r(\gamma^{K}_{ij})].
\label{eq:fullmodelprobability}
\end{aligned}
\end{equation}
Also, the binary cross-entropy is used as training objection, which is the same as the advanced DocRE model~\cite{yao-etal-2019-docred}.

\section{Experiments}
\label{sec4}
\subsection{Data set and Setup}
\label{sec4-1}
\begin{table}[ht!]
\centering
\begin{tabular}{lr}
\hline
Hyperparameter             & Value      \\ \hline
Batch Size                 & 12         \\
Optimizer                  & AdamW      \\
Learning Rate              & 1e-3       \\
Activation Function        & ReLU       \\
Word Embedding Size        & 100        \\
Entity Type Embedding Size & 20         \\
Coreference Embedding Size & 20         \\
Encoder Hidden Size        & 128        \\
Dropout                    & 0.5        \\
Layers of GCN              & 2          \\
Weight Decay               & 0.0001     \\ 
Device                     & GTX 1080Ti \\ \hline
\end{tabular}
\caption{Settings for DRN.}
\label{table:glove_hyperparameters} 
\end{table}
The proposed methods were evaluated on a large-scale human-annotated dataset for document-level relation extraction~\cite{yao-etal-2019-docred}. 
DocRED contains 3,053 documents for the training set, 1,000 documents for the development set, and 1,000 documents for the test set, totally with 132,375 entities, 56,354 relational facts, and 96 relation types. 
More than 40\% of the relational facts require reading and reasoning over multiple sentences.
For more detailed statistics about DocRED, we recommend readers to refer to the original paper~\cite{yao-etal-2019-docred}.

Following settings of~\citeauthor{yao-etal-2019-docred}'s work, we used the GloVe embedding (100d) and BiLSTM (128d) as word embedding and encoder.
The number of the reasoning path for each task is set to 3.
The learning rate was set to 1e-3 and we trained the model using AdamW~\cite{adamw2019} as the optimizer with weight decay 0.0001 under Pytorch~\cite{paszke2017pytorch}.
For the BERT representations, we used uncased BERT-Based model (768d) as the encoder and the learning rate was set to $1e^{-5}$.
For evaluation, we used F1 and Ign F1 as the evaluation metrics. 
Ign F1 denotes F1 score excluding relational facts shared by the training and development/test sets.
In particular, the predicted results were ranked by their confidence and traverse this list from top to bottom by $F1$ score on development set, and the score value corresponding to the maximum $F1$ is picked as threshold $\theta$.
The hyper-parameter for the number of reasoning paths was tuned based on the development set.
In addition, the results on the test set were evaluated through CodaLab\footnote{\url{https://competitions.codalab.org/competitions/20717}}.
Once a model is trained, we get the confidence scores for every triple example (subject,object,relation) as Eq.(\ref{eq:modelprobabilitymax}). 
We rank the predicted results by their confidence and traverse this list from top to bottom by F1 score on development set, the score value corresponding to the maximum F1 is picked as threshold $\theta$. 
This threshold is used to control the number of extracted relational facts on the test set.

\subsection{Baseline Systems}
\label{sec4-2}
We reported the results of the recent graph-based DocRE methods as the comparison systems: GAT~\cite{Velickovic2018GraphAN}, GCNN~\cite{sahu-etal-2019-inter}, EoG ~\cite{Christopoulou2019ConnectingTD}, AGGCN~\cite{guo-etal-2019-attention}, \textbf{LSR}~\cite{Nan2020ReasoningWL}, \textbf{GAIN}~\cite{zeng-etal-2020-double}, and \textbf{HeterGASN-Rec}\cite{docred-rec}.
Moreover, pre-trained models like \textbf{BERT}~\cite{devlin-etal-2019-bert} has been shown impressive result on the DocRE task.
Therefore, we also reported state-of-the-art graph-based DocRE models with pre-trained BERT$_{base}$ model, including 
\textbf{Two-Phase+BERT$_{base}$}~\cite{Wang2019FinetuneBF}, 
\textbf{LSR+BERT$_{base}$}~\cite{Nan2020ReasoningWL},
\textbf{GAIN+BERT$_{base}$}~\cite{zeng-etal-2020-double}, 
\textbf{HeterGASN-Rec+BERT$_{base}$}~\cite{docred-rec}, and \textbf{ATLOP-BERT$_{base}$}~\cite{zhou2021atlop}.

\subsection{Main Results}
\label{sec4-3}
Table \ref{tab:mianresult} presents the detailed results on the development set and the test set for the DocRE dataset. 
First, the proposed DRN model significantly outperformed the existing graph-based DocRE systems. 
Second, the proposed DRN model was superior to all the existing graph-based DocRE systems on the test set, validating that modeling reasoning discriminatively is more beneficial to DocRE than the original universal neural network way.
Meanwhile, it also outperformed the best HeterGSAN-Rec model by 1.10 points in terms of F1, validating the effectiveness of our discriminative reasoning method.
Third, for the comparisons with a pre-trained language model (BERT$_{base}$), F1 scores of the proposed DRN+BERT$_{base}$ model was higher than that of the existing graph-based DocRE ATLOP+BERT model systems with BERT$_{base}$ on the test set.
In particular, our method (F1 61.37) was superior to the existing best ATLOP+BERT model (F1 61.30) in terms of F1, which is a new state-of-the-art result on the DocRE dataset.
\begin{table}[h]
\begin{center}
\scalebox{.74}{
\begin{tabular}{l|l|l|l|l}
\hline 
\multicolumn{1}{c|}{\multirow{2}{*}{Methods}} & \multicolumn{2}{c|}{Dev} & \multicolumn{2}{c}{Test} \\ \cline{2-5} 
                       \multicolumn{1}{c|}{}                       & Ign F1        & F1       & Ign F1        & F1        \\ \hline
\multicolumn{5}{c}{\textit{Existing DocRE Systems}}                                                        \\ \hline 
    GCNN$^\dag$          & 46.22 & 51.52 & 49.59 & 51.62         \\ 
    EoG$^\dag$  & 45.94 & 52.15 & 49.48 & 51.82        \\ 
    GAT$^\dag$         & 45.17 & 51.44 & 47.36 & 49.51          \\ 
    AGGCN$^\dag$      & 46.29 & 52.47 & 48.89 & 51.45        \\ 
    LSR$^*$         & 48.82 & 55.17 & 52.15 & 54.18        \\ 
    GAIN$^*$      & 53.05 & 55.29 & 52.66 & 55.08 \\
    HeterGSAN-Rec$^*$   & 54.27 & 56.22 & 53.27 & 55.23 \\ \cdashline{1-5}
    BERT$_{base}^*$               & -     & 54.16 & -     & 53.20      \\ 
    Two-Phase BERT$_{base}^*$     & -     & 54.42 & -     & 53.92          \\ 
    LSR+BERT$_{base}^*$           & 52.43 & 59.00 & 56.97 & 59.05        \\ 
    GAIN+BERT$_{base}^*$       & 59.14 & 61.22 & 59.00 & 61.24        \\ 
    HeterGSAN-Rec+BERT$_{base}^*$         & 58.13 & 60.18 & 57.12 & 59.45       \\ 
    ATLOP-BERT$_{base}^*$             & 59.22 & 61.09 & 59.31 & 61.30     \\\hline
    \multicolumn{5}{c}{\textit{Our DocRE Systems}}   \\  \cline{1-5}
    \textbf{DRN}   & 54.61 & 56.49 & 54.35 & 56.33     \\ \cdashline{1-5}
   \textbf{DRN}+BERT$_{base}$ & 59.33 & 61.39 & 59.15 & 61.37   \\ \hline 
\end{tabular}}
\end{center}
\caption{Results on the development set and the test set. Results with $*$ are reported in their original papers. Results with $\dag$ are reported in \cite{Nan2020ReasoningWL}. Bold results indicate the best performance of the current method.}
\label{tab:mianresult} 
\end{table}

\subsection{Evaluating Hyper-parameter $K$ for The Number of Reasoning Paths}
\begin{figure*}[ht!]
\subfigure{
	\begin{minipage}[t]{0.48\textwidth}
			\centering
			\pgfplotsset{height=5.2cm,width=7.5cm,compat=1.14,every axis/.append style={thick}}
			\begin{tikzpicture}
                \begin{axis}[
                	ybar=5pt,
                	tick align=inside,
                	bar width=9pt,
                	enlargelimits=0.15,
                	legend style={at={(0.51,0.95)},
                		anchor=north,legend columns=1},
                	ylabel={Percentage},
                	nodes near coords,
                	symbolic x coords={IR,LR,CR},
                	xtick=data,
                	tick align=inside,
                ]
                \addplot[white!30!brown,fill=white!30!brown]
                	coordinates {(IR,19.12)(LR,19.17)(CR,61.71)};
                \addplot[white!40!purple,fill=white!40!purple]
                	coordinates {(IR,47.58)(LR,13.91)(CR,38.51)};
                
                \legend{\small{All entity pairs}, \small{Ones with relation}}
                \end{axis}
            \end{tikzpicture}
			\centerline{(a)}
	\end{minipage}}
		\subfigure{
		\begin{minipage}[t]{0.48\textwidth}
			\centering
			\setlength{\abovecaptionskip}{0pt}
			\pgfplotsset{height=5.2cm,width=7.5cm,compat=1.14,every axis/.append style={thick}}
            \begin{tikzpicture}
            \begin{axis}[
            	x tick label style={
            		/pgf/number format/1000 sep=},
            	ylabel=F1 Score,
            	enlargelimits=0.15,
            	legend style={at={(0.66,0.95)},
            		anchor=north,legend columns=3},
            	ybar,
            	tick align=inside,
            	bar width=8pt,
            	symbolic x coords={IR,LR,CR},
            	xtick=data,
            ]
            \addplot[white!60!orange,fill=white!60!orange]
            	coordinates {(IR,63.91) (LR,38.72) (CR,47.85)};
            \addplot[white!60!blue,fill=white!60!blue]
            	coordinates {(IR,64.25) (LR,41.80) (CR,49.30)};
            \addplot[white!60!red,fill=white!60!red]
            	coordinates {(IR,64.63) (LR,42.27) (CR,49.56)};
            
            \legend{\small{GAIN},\small{Rec},\small{DRN}}
            \end{axis}
            \end{tikzpicture}
			\centerline{(b)}
	\end{minipage}}
	
	\caption{(a) The statistical result of different reasoning task. (b) The performance of different reasoning task.}
	\label{fig:taskperformance}
\end{figure*}
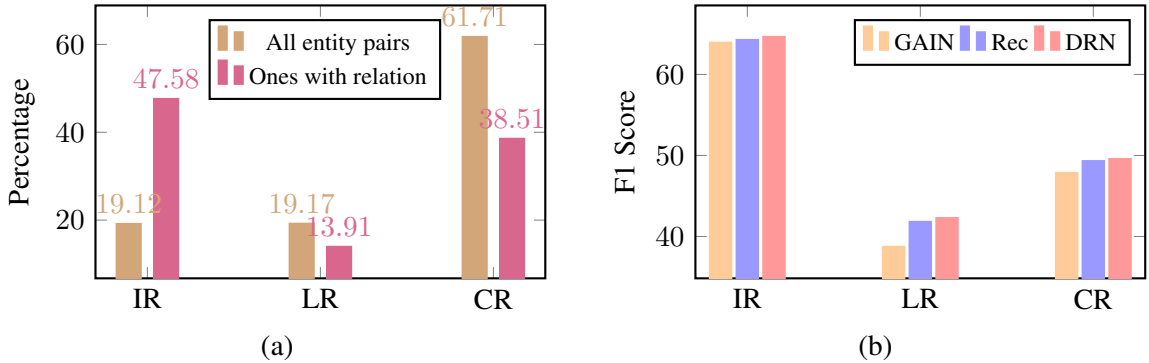
\label{sec4-4}
\begin{table}[h]
\centering
\scalebox{.9}{
\begin{tabular}{c|cccc|c}
\hline
\multirow{2}{*}{K} & \multicolumn{2}{c}{Dev Set} & \multicolumn{2}{c|}{Test Set} & \multirow{2}{*}{\begin{tabular}[c]{@{}c@{}}Cover \\ (\%)\end{tabular}} \\ \cline{2-5}
                   & Ign F1        & F1          & Ign F1         & F1           &                        \\ \hline
1                  & 54.04         & 55.94       & 53.83          & 55.81        &  63.05                      \\
2                  & 54.63         & 56.47       & 54.12          & 56.07        &  82.17                      \\
3                  & 54.61         & 56.49       & 54.35          & 56.33        &  90.40                      \\
4                  & 54.52         & 56.34       & 54.06          & 55.93        &  95.22                      \\
\textgreater{}4    & 54.31         & 56.25       & 53.97          & 55.84        &  100                    \\ \hline
\end{tabular}
}
\caption{The effect of the number of reasoning paths $K$ for the proposed DRN model.}
\label{table:max_path} 
\end{table}
To evaluate the effect of the number of reasoning path $K$ in Eq.\ref{eq:multifullprobability}, we reported the results for the different number of reasoning path $K$, as shown in Table \ref{table:max_path}.
When $K$ increased from 1 to 3, F1 scores of the proposed DRN model gradually improved from 55.81 to 56.33 on the test set and the percentage of covered reasoning paths reaches 90.40\%.
As the hyper-parameter $K$ continues to increase, F1 scores began to drop on the dev and test sets.
On the one hand, the reason may be that the reasoning information provided by too many reasoning paths is duplicated, even noises in the remaining 9.60\% reasoning paths.
On the other hand, the hyper-parameter $K$=3 can make the proposed DRN gain the highest F1 score on the dev and test sets.
Therefore, we set the hyper-parameter $K$ to three in our main results in Table~\ref{tab:mianresult}.

\subsection{Ablation Experiments}
In the proposed DRN model, we model different reasoning tasks discriminatively using HGCRep and DLCRep, and we choose the highest scores as the final results.
Instead of using the discriminative reasoning framework, previous work averaged the mention representation (HGCRep or DLCRep) to get the entity representation and concatenate the two entity representation to classify the relation, which we denote as Uniform model.
Table \ref{table:ablation} shows ablation experiments of the framework and different reasoning context on the test set. 
It is noted that Uniform model with the discriminative reasoning framework is our DRN model.
First, the DocRE models benefit from our discriminative reasoning framework no matter what the reasoning context is used.
Specially, the F1 score of the model with the framework was averagely 1.21 points superior to the Uniform model on the test set no matter what context representation is used, which illustrated the effectiveness of the framework.
\begin{table}[h]
\begin{center}
\scalebox{.83}{
\begin{tabular}{l|cl|cl|c}
\hline
\multicolumn{1}{c|}{Model} & \multicolumn{2}{c|}{\begin{tabular}[c]{@{}c@{}}without \\ framework\end{tabular}} & \multicolumn{2}{c|}{\begin{tabular}[c]{@{}c@{}}with \\ framework\end{tabular}} & Delta                      \\ \cline{2-6} 
                           & \multicolumn{1}{l}{Ign F1}                         & F1                           & \multicolumn{1}{l}{Ign F1}                       & F1                          & \multicolumn{1}{l}{Ign F1} \\ \hline
Uniform                    & 53.68                                              & 55.79                        & 54.35                                            & 56.33                       & +0.83                      \\
-DLCReps                   & 51.82                                              & 53.83                        & 52.96                                            & 55.01                       & +1.33                      \\
-HGCReps                   & 51.21                                              & 53.36                        & 52.35                                            & 54.13                       & +0.97                      \\
-Both                      & 44.73                                              & 51.06                        & 50.68                                            & 52.78                       & +1.71                      \\ \hline
\end{tabular}}
\caption{Ablation experiments.}
\label{table:ablation}
\end{center}
\end{table}
Second, when we gradually remove DLCRep and HGCRep from the Uniform and the proposed DRN model, both of the model's performance drops.
Specially, F1 scores of DRN without DLCRep dropped by 1.32 while F1 scores of DRN without HGCRep dropped by 2.20 respectively. 
This indicates that both DLCRep and HGCRep play an important role in capturing the information of nodes on the reasoning paths. 
When removing both of DLCPeps and HGCReps from the DRN model, the model was degraded to the BiLSTM model with our discriminative reasoning framework. 
Obviously, F1 scores drastically decreased on the test sets, confirming the necessity of learning DLCRep and HGCRep for modeling reasoning discriminatively.

\subsection{Analysis of the Reasoning Tasks}
\begin{figure*}[h!]
  \centering
  \includegraphics[scale=0.89]{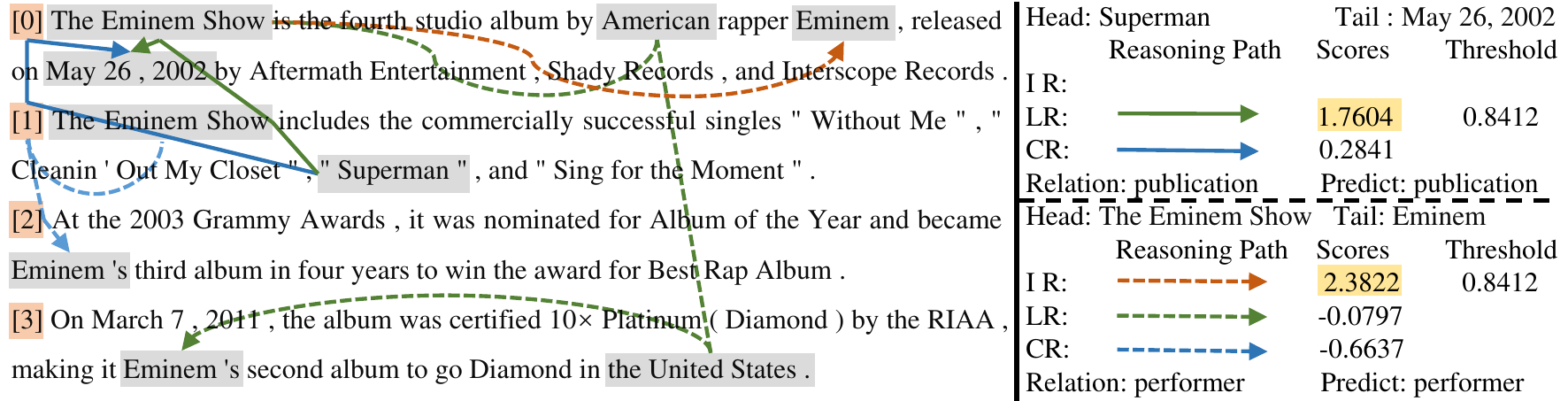}
  \caption{Case study.}
  \label{fig:casestudy}
\end{figure*}
In this section, we first showed the percent of all entity pairs (396,790) and entity pair with relation  (12,332) on the dev set selected for three defined reasoning tasks through max operation in Eq.\eqref{eq:modelprobabilitymax}, as shown in Figure~\ref{fig:taskperformance}(a).
For example, IR, LR, and CR are the intra-Sentence reasoning task, the logical reasoning task, and the coreference reasoning task, respectively.
The percentages of IR, LR, and CR which is selected for all the entity pair are 19.12\%, 19.17\%, and 61.71\% for all entity pairs, respectively.
This indicates that our defined three reasoning skills can completely cover all entity pairs regardless of whether these entity pairs have relationships or not.
Also, the percentages of IR, LR, and CR are 47.58\%, 13.91\%, and 38.51\% for entity pairs with relation, respectively.
This is consistent with the statistical result in the \citeauthor{yao-etal-2019-docred}'s work that more than 40.7\% relational facts can only be extracted from multiple sentences, validating that our method can model different reasoning skills discriminatively on the DocRE dataset.

Moreover, Figure \ref{fig:taskperformance}(b) showed the results of HerterGSAN-Rec (abbreviated as Rec), GAIN, and our DRN models on three different reasoning tasks.
As seen, F1 scores of the proposed DRN model are higher than that of Rec and GAIN models over all three tasks.
This means that modeling reasoning types explicitly can effectively advance the DocRE.
For all DocRE models, F1 scores of LR task and CR task were far inferior to that of IR task, which is consistent with the intuitive perception that the inter-sentence reasoning is more difficult than the intra-sentence reasoning.

\subsection{Analysis of the Reasoning Type}
\begin{table}[h]
\centering
\scalebox{.8}{
\begin{tabular}{cccccc}
\hline
\multicolumn{6}{c}{Confusion Matrix}                                                                                                                                   \\ \hline
                                              & \multicolumn{1}{c|}{}      & \multicolumn{3}{c|}{Truth}                                                        &       \\ \cline{3-6} 
                                              & \multicolumn{1}{c|}{}      & IR                         & LR                      & \multicolumn{1}{c|}{CR}    & Total \\ \hline
\multicolumn{1}{c|}{\multirow{4}{*}{Predict}} & \multicolumn{1}{c|}{IR}    & 321                        & 36                      & \multicolumn{1}{c|}{12}    & 369   \\
\multicolumn{1}{c|}{}                         & \multicolumn{1}{c|}{LR}    & 25                         & 88                      & \multicolumn{1}{c|}{29}    & 142    \\
\multicolumn{1}{c|}{}                         & \multicolumn{1}{c|}{CR}    & 88                         & 129                     & \multicolumn{1}{c|}{188}   & 405   \\ \cline{2-6} 
\multicolumn{1}{c|}{}                         & \multicolumn{1}{c|}{Total} & 434                        & 253                     & \multicolumn{1}{c|}{229}   &       \\ \hline \hline
\multicolumn{6}{c}{Metric}                                                                                                                                             \\ \hline
\multicolumn{2}{c|}{\multirow{2}{*}{}}                                   & \multicolumn{1}{c|}{IR}    & \multicolumn{1}{c|}{LR} & \multicolumn{1}{c|}{CR}    &       \\ \cline{1-6} 
\multicolumn{2}{c|}{}                                               F1 Scores       & 79.95                      & 38.77                   & \multicolumn{1}{c|}{44.87} &       \\ \cline{3-6} 
\hline
\end{tabular}
}
\caption{Confusion matrix of different reasoning types.}
\label{table:reasoningtype} 
\end{table}
To further show the selected different reasoning types in Eq.\eqref{eq:modelprobabilitymax}, we randomly sampled 72 documents from the dev set which contain 916 relation instances, and we ask three human to annotate the reasoning types of all the entity pairs with relation in the sampled document according to three defined reasoning types, including the intra-sentence reasoning, the logical reasoning, and the coreference reasoning (The annotation data can be found in \url{https://github.com/xwjim/DRN}).
Table \ref{table:reasoningtype} shows the number and F1 scores of each selected reasoning types on the sampled 72 documents.
As seen, F1 scores of IR, LR, and CR are 79.95\%, 38.77\%, and 44.87\%, respectively, indicating that modeling reasoning discriminatively is working during selecting of reasoning paths in Eq.\eqref{eq:modelprobabilitymax}.
Also, our method is the capacity of recognizing not only the intra-sentence reasoning but also the intra-sentence reasoning.
In addition, there is a certain percentage of the mistakenly selected reasoning types, indicating that our method may have more room for improvement in the future.

\subsection{Case Study}
Figure \ref{fig:casestudy} shows the relation classification about two entity pairs for our DRN model. 
For the first entity pair \{``\textit{Superman}"\} and \{``\textit{May 26,2002}"\}, there are reasoning paths for Task2 and Task3, and their scores are 1.7604, and 0.2841,respectively
As a result, Task2 was used to predict the relation ``\{\textit{publication date}\}" between \{``\textit{Superman}"\} and \{``\textit{May 26,2002}"\} correctly.
Meanwhile, the selection of Task2 is consistent with the ground-truth logical reasoning type.
Moreover, the above reasoning processing is also similar to the entity pair \{``\textit{The Eminem show}"\} and \{``\textit{Eminem}"\} with three reasoning types.

\section{Related Work}
Early research efforts on relation extraction concentrate on predicting the relation between two entities with a sentence~\cite{zeng-etal-2014-relation,zeng-etal-2015-distant,wang-etal-2016-relation,sorokin-gurevych-2017-context,reinclass,Song_2019,wei2019novel}. 
These approaches do not consider interactions across mentions and ignore relations expressed across sentence boundaries. The semantics of a document context is coherent and a part of relation can only be extracted among sentences. 

However, as large amounts of relationships are expressed by multiple sentences, recent work starts to explore document-level relation extraction.
People begin to consider the relation between disease and chemicals in the entire document of biomedical domain \cite{quirk-poon-2017-distant,DBLP:journals/corr/abs-1810-05102,zhang-etal-2018-graph,Christopoulou2019ConnectingTD,zhu-etal-2019-graph}.
A large-scale general-purpose dataset for DocRE constructed from Wikipedia articles has been proposed in \cite{yao-etal-2019-docred}, which has advanced the DocRE a lot.
Most approaches on DocRE are based on document graphs, which were introduced by \citeauthor{quirk-poon-2017-distant}. Specifically, they use words as nodes and construct a homogenous graph using syntax parsing tools and a graph neural network is used to capture the document information. 
This document graph provides a unified way of extracting the features for entity pairs. 
Later work extends the idea by improving neural architectures~\cite{DBLP:journals/tacl/PengPQTY17,verga-etal-2018-simultaneously,DBLP:journals/corr/abs-1810-05102} or adding more types of edges~\cite{Christopoulou2019ConnectingTD}.
In the \citeauthor{Christopoulou2019ConnectingTD}'s work, the author construct the graph which contains different granularities (sentence, mention, entity) through co-occurrence and heuristic rule to model the graph without external tools.
More recent most of the approach \cite{Christopoulou2019ConnectingTD,zeng-etal-2020-double,docred-rec} constructs heterogeneous graph through co-occurrence and heuristic rule to model the graph without external tools.
In the \cite{zeng-etal-2020-double} constructed double graphs in different granularity to capture document-aware features and the interaction between entities.
In the \cite{docred-rec} introduced a reconstructor to reconstruct the path in the graph to guide the model to learning a good node representation.
Other attempts focus on the multi-entity and multi-label problems ~\cite{zhou2021atlop}. \citeauthor{zhou2021atlop} proposed two techniques to solve the problems, adaptive thresholding and localized context pooling. 
\section{Conclusion}
In this paper, we propose a novel discriminative reasoning framework to consider different reasoning types explicitly. 
We use meta-path strategy to extract the reasoning path for different reasoning types.
Based on the framework, we propose a Discriminative Reasoning Network~(DRN), in which we use both the heterogeneous graph context and the document-level context to represent different reasoning paths. 
The ablation study validates the effectiveness of our discriminative framework and different modules on the large-scale human-annotated DocRE dataset.
In particular, our method archives a new state-of-the-art performance on the DocRE dataset.
In the future, we will explore more diverse structure information~\cite{AAAI1816060,9097389,Cohen2020Scalable} from the input document for the discriminative reasoning framework, and apply the proposed approach to other NLP tasks~\cite{zhang-etal-2020-two,chen-etal-2020-aspect,zhang2020sg}.  

\section*{Acknowledgments}
We are grateful to the anonymous reviewers, area chair and Program Committee for their insightful comments and suggestions.
The corresponding authors are Kehai Chen and Tiejun Zhao.
The work of this paper is funded by the project of National Key Research and Development Program of China (No. 2020AAA0108000).

\bibliographystyle{acl_natbib}
\bibliography{acl2021}

\end{document}